\documentclass[10pt,twocolumn,letterpaper]{article}

\usepackage{cvpr}              

\usepackage{graphicx}
\usepackage{amsmath}
\usepackage{amssymb}
\usepackage{booktabs}
\usepackage{multirow}
\usepackage{physics}
\usepackage{algorithm,algorithmic}
\usepackage{enumitem}

\usepackage{epsfig}
\usepackage{multirow}
\usepackage{caption}
\usepackage{rotating}

\usepackage[pagebackref,breaklinks,colorlinks]{hyperref}

\usepackage[capitalize]{cleveref}
\crefname{section}{Sec.}{Secs.}
\Crefname{section}{Section}{Sections}
\Crefname{table}{Table}{Tables}
\crefname{table}{Tab.}{Tabs.}

\def\cvprPaperID{2879} 
\def\confName{CVPR}
\def\confYear{2022}

\def\framework{{\textcolor{black}{X-Learner}}}
\def\dilationstage{{\textcolor{black}{Expansion Stage}}}
\def\squeezestage{{\textcolor{black}{Squeeze Stage}}}
\def\translayer{{\textcolor{black}{reconciliation layer}}}

\begin{document}

\title{\framework{}: Learning Cross Sources and Tasks for \\ Universal Visual Representation}

\author{Yinan He$^{1}$\thanks{} \quad Gengshi Huang$^{1*}$ \quad Siyu Chen$^{1*}$ \quad Jianing Teng$^{1*}$ \\
Wang Kun$^{1}$ \quad Zhenfei Yin$^{1}$ \quad Lu Sheng$^{2}$ \quad Ziwei Liu$^{3}$ \quad Yu Qiao$^{4}$ \quad Jing Shao$^{1}$\thanks{} \\ 
$^{1}$SenseTime Research \quad $^{2}$College of Software, Beihang University  \\ \quad $^{3}$S-Lab, Nanyang Technological University  \quad $^{4}$Shanghai AI Laboratory \\
{\tt\small \{heyinan, huanggengshi, chensiyu, tengjianing, wangkun, yinzhenfei, shaojing\}@senseauto.com}\\
{\tt\small lsheng@buaa.edu.cn} \quad 
{\tt\small ziwei.liu@ntu.edu.sg} \quad
{\tt\small qiaoyu@pjlab.org.cn}
}

\vspace{-1cm}
\twocolumn[{%
\maketitle
\vspace{-20pt}
\begin{figure}[H]
    \hsize=\textwidth
    \centering
    \includegraphics[width=2.1\linewidth]{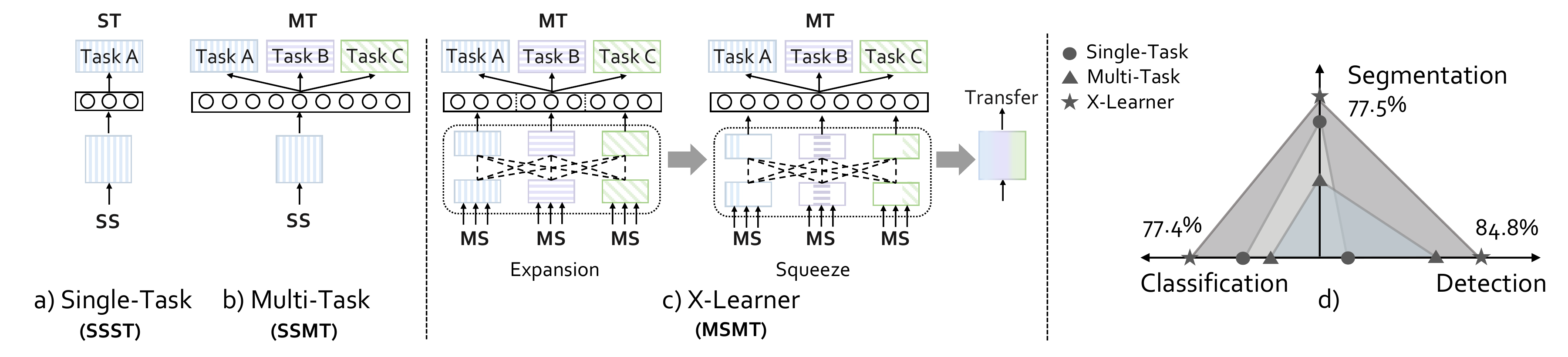}
    \vspace{-0.5cm}
    \captionof{figure}{a) Single-Source Single-Task; b) Single-Source Multi-Task; c) \framework{}: Multi-Source Multi-Task; d) Our proposed X-Learner achieves the best performances in Classification (average linear probe results across 10 classification datasets), Detection (Pascal VOC Detection~\cite{Everingham10VOC}) and Segmentation (Pascal VOC Semantic Segmentation~\cite{Everingham10VOC}).}
    \label{fig:figure1}
\end{figure}
}]
{
  \renewcommand{\thefootnote}%
    {\fnsymbol{footnote}}
  \footnotetext[1]{Equal contribution.}
  \footnotetext[2]{Corresponding author.}
}


\begin{abstract}
In computer vision, pre-training models based on large-scale supervised learning have been proven effective over the past few years. However, existing works mostly focus on learning from individual task with single data source (\emph{e.g.}, ImageNet for classification or COCO for detection). This restricted form limits their generalizability and usability due to the lack of vast semantic information from various tasks and data sources. Here, we demonstrate that jointly learning from heterogeneous tasks and multiple data sources contributes to universal visual representation, leading to better transferring results of various downstream tasks. Thus, learning how to bridge the gaps among different tasks and data sources is the key, but it still remains an open question.
In this work, we propose a representation learning framework called \textbf{\framework{}}, which learns the universal feature of multiple vision tasks supervised by various sources, with expansion and squeeze stage:
\textbf{1) \dilationstage{}:} X-Learner learns the task-specific feature to alleviate task interference and enrich the representation by reconciliation layer.
\textbf{2) \squeezestage{}:} X-Learner condenses the model to a reasonable size and learns the universal and generalizable representation for various tasks transferring.
%
Extensive experiments demonstrate that X-Learner achieves strong performance on different tasks without extra annotations, modalities and computational costs compared to existing representation learning methods.
%
Notably, a single X-Learner model shows remarkable gains of 3.0\%, 3.3\% and 1.8\% over current pre-trained models on 12 downstream datasets for classification, object detection and semantic segmentation.

\end{abstract}

\section{Introduction}
\label{sec:intro}



Substantial advances have been achieved in visual representation learning, such as those based on curated large-scale image datasets with supervised~\cite{kolesnikov2020BiT,sun2017revisiting}, weakly-supervised~\cite{joulin2016learning,mahajan2018exploring}, semi-supervised~\cite{yalniz2019billion,yan2020clusterfit}, as well as self-supervised~\cite{mocov1,chen2020mocov2,simsiam,caron2020unsupervised,byol} pre-training.
These visual representations show promising abilities in improving the performance on downstream tasks.

Among these pre-training techniques,  supervised pre-training is widely adopted for its clear objective and steady training process. Nevertheless, existing works in this direction only consider individual upstream task\footnote{To avoid ambiguity, we refer to a \textit{task} as a general vision problem such as classification, detection or segmentation, and a \textit{source} as a specific dataset or context within a certain \textit{task}.} (e.g., classification or detection) and most of them solely utilize one single data source (e.g., ImageNet~\cite{deng2009imagenet} or COCO~\cite{lin2014microsoft}).
We argue this single-source single-task (SSST, \cref{fig:figure1} (a)) paradigm has several drawbacks:
1) The learned representation in SSST is specialized for one given task and is likely to have inferior performance on other tasks~\cite{sermanet2013overfeat,mensink2021factors,ghiasi2021multi,rethink1,shao2019objects365}.
2) It misses the potentials of a more robust representation by integrating characteristic semantic information from different tasks.  
Intuitively, we can opt to a simple hard-sharing method, i.e. single-source, multi-task (SSMT) paradigm, as described in~\cref{fig:figure1} (b), by building many heads, each of which is specific for one task ~\cite{sermanet2013overfeat,han2017heterogeneous}.
However, this over-simplified algorithm usually encounters task interference~\cite{zhao2018modulation,maninis2019attentive}, especially for heterogeneous tasks, leading to a significant drop in performance.
Besides, it requires the same image with a variety of labels~\cite{taskonomy,zamir2020robust}, which is not scalable easily due to the high annotation cost.
A recent self-training work~\cite{ghiasi2021multi} attempts to create a pseudo multi-task dataset to alleviate the data-scarcity issue of multi-task learning, which follows a similar spirit to other SSMT works.

In light of issues with previous settings, we focus on utilizing numerous data sources of multiple tasks to learn a universal visual representation which should transfer well to various downstream tasks like classification, object detection and semantic segmentation. 
To leverage cross-source, cross-task information and mitigate undesired task interference,
we propose a new pre-training paradigm~\textit{\textbf{X-Learner}}, as shown in~\cref{fig:figure1} (c). 
The X-Learner contains two dedicated stages:
\textbf{1) \dilationstage{}:} It first trains a set of sub-backbones, each of which specifically exploits one task enriched with multiple sources. It then joins together these sub-backbones and combine their representational knowledge via our proposed \emph{reconciliation layer}, forming an expanded backbone with enhanced modeling capacity.
\textbf{2) \squeezestage{}}: Given the expanded backbone, this stage reduces the model complexity back to sub-backbone level and produces a unified and compact multi-task-aware representation.
This new paradigm has two main advantages:
\textbf{1)} It can effectively consolidate diverse knowledge from our new multi-source multi-task learning and avoid task conflicts. The resulting representation generalizes well to different types of tasks simultaneously.
\textbf{2)} Compared to traditional multi-task methods, it is highly extensible with new tasks and sources, since we only require data sources annotated with single-task labels.

Our contributions are summarized as follows:

\begin{itemize}[leftmargin=*]

\item We propose a new \textbf{multi-source multi-task learning} setting that only requires single-task label per datum, and is highly scalable with more tasks and sources without requiring any extra annotation effort.

\item We present \textbf{\framework{}}, a general framework for learning a universal representation from supervised multi-source multi-task learning, with \dilationstage{} and \squeezestage{}. Task interference can be well mitigated by \dilationstage{},
while a compact and generalizable model is produced by \squeezestage{}.
With \framework{}, heterogeneous tasks can be jointly learned, and the resulting single model renders a universal visual representation suitable for various tasks.

\item We show the \textbf{strong transfer ability} of feature representations learned by our \framework{}. In terms of transfer learning performance, multi-source multi-task learning with our two-stage design outperforms traditional supervised single/multi-task training, self-supervised learning and self-training methods. As illustrated in \cref{fig:figure1} (d), a model pre-trained with \framework{} exhibits significant gains (3.0\%, 3.3\% and 1.8\%) over the ImageNet supervised counterpart on downstream image classification, object detection and semantic segmentation. 

\item We offer \textbf{several new insights} into representation learning and the framework design for multi-task and multi-source learning through extensive experiments. 

\end{itemize}

\section{Related Work}
\label{sec:related-work}

\noindent\textbf{Visual Representation Learning.}
Significant progress has been made in the field of visual representation learning, including unsupervised method~\cite{chen2020simple,ssl2,ssl3,mocov1,chen2020mocov2,clip},  supervised training~\cite{kolesnikov2020BiT,sun2017revisiting}, weakly-supervised learning~\cite{joulin2016learning,mahajan2018exploring}, and semi-supervised learning~\cite{yalniz2019billion,yan2020clusterfit}.
A large quantity of prior works use supervised datasets, including ImageNet1k~\cite{supervised1}, ImageNet-21K~\cite{ridnik2021imagenet}, IG-3.5B-17k~\cite{mahajan2018exploring} and JFT~\cite{kolesnikov2020BiT}, for learning visual representations.
In supervised pre-training, labeled training data provide significant improvement for transfer performance in the same task as the one for which the data are annotated. However, the ability of transferring across different tasks is not good enough~\cite{rethink2}. 
In unsupervised learning, \cite{clip} focuses on multi-modal vision language pre-training to achieve strong performances in classification, but not do well in other visual tasks like detection~\cite{gu2021zero}.
In order to obtain uniformly high transfer performance on diverse task types, it is important to improve the task diversity of training data, justifying the necessity of multi-task pre-training.

\begin{figure*}[t]
    \centering
     \includegraphics[width=1.0\linewidth]{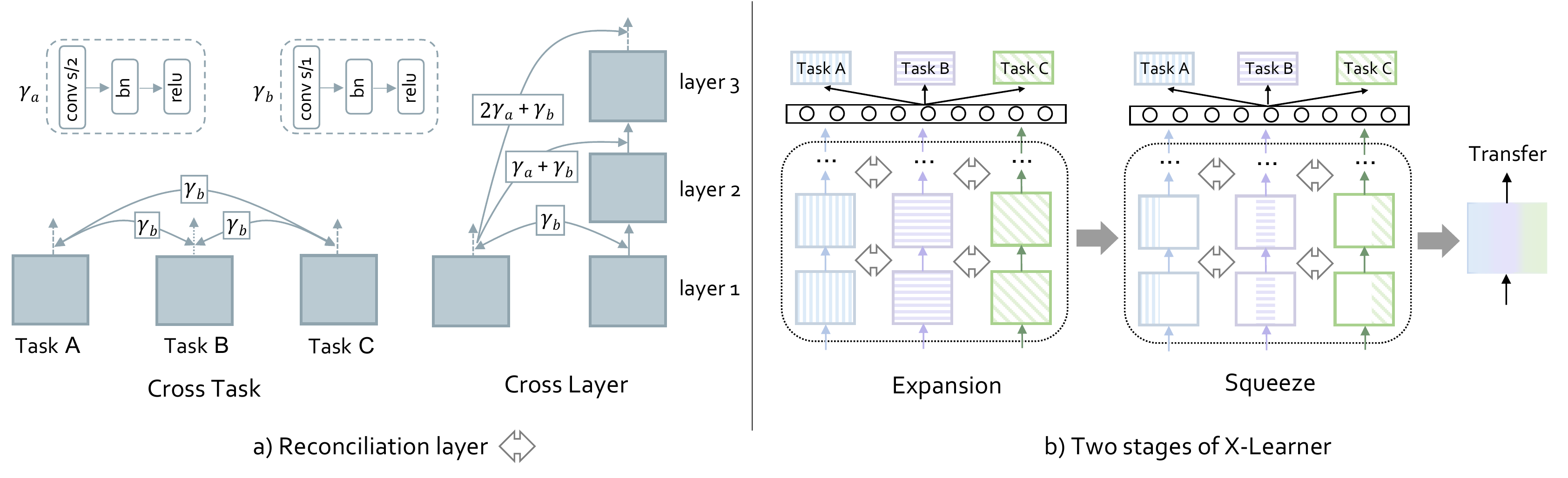}
     \caption{\textbf{Structure of \framework{}.}
    a) illustrates how \translayer{}s make the features from different tasks interact with each other. We use $\gamma$ to represent the reconciliation layer. We present two typical ways of connection by \translayer{}: cross different tasks and cross multiple layers; b) Features for different tasks are learned in \dilationstage{} and unified in \squeezestage{}. After the two stages, X-Learner obtains a general representation for transferring to downstream tasks.}
     \label{fig:stage3-framework}
  \end{figure*}

\vspace{2mm}
\noindent\textbf{Multi-Task Learning.}
There has been substantial interest in multi-task learning~\cite{taskonomy,liu2019endattention,caruana1997multitask,bilen2017universal,rebuffi2017learning,guo2019depthwise, zhao2020object,Towards_universal,Simple_multi_dataset_detection} in the community. A common practice for multi-task learning is to share the hidden layers of a backbone model across different tasks, which is called ``hard-sharing'' in the literature. However, such sharing is not always beneficial, in many times hurting performance~\cite{yu2020gradientsurg, yang2021improving, wang2020gradientvac, guo2019depthwise}. To alleviate this, there are several lines of works to solve the problem in different ways. One of them is the use of a split architecture with parallel backbones for different tasks~\cite{misra2016cross, liu2019endattention, gao2019nddr}. \cite{misra2016cross} proposes a cross-stitch module, which intelligently combines task-specific networks, avoiding the need to brute-force search through numerous architectures. 
Another line of works is improving optimization during learning~\cite{yu2020gradientsurg, yang2021improving, wang2020gradientvac, li2020knowledge}. For example, \cite{yu2020gradientsurg} mitigates gradient interference by altering the gradients directly, i.e., performing ``gradient surgery''. \cite{wang2020gradientvac} addresses interference by de-conflicting gradients via projection. \cite{li2020knowledge,li2021representation} use distillation to avoid interference, but they are limited to a retrained setting, either single-task multi-source or single-source multi-task.
Other works attempt to develop systematic techniques to determine which tasks should be trained together in a multi-task neural network to avoid harmful conflicts between non-affinitive tasks~\cite{fifty2021efficientlygrouping, baxter2000model, ben2003exploiting, kumar2012learning, achille2021information}. 
These methods perform multi-task learning to improve the performances of tasks involved, but they are not concerned with the transfer performance on downstream tasks. 
~\cite{polyvit} applies vision transformer on multiple modalities and achieves impressive performance. For the image modality, it deals with the classification task only, and learns in a simple hard-sharing way. The problem of multi-task learning remains. A recent work~\cite{ghiasi2021multi} turns to semi-supervised learning and constructs cross-task pseudo labels with task-specific teachers, creating a complete multi-task dataset for pre-training. Yet it only considers the single-source setting, and its student training still follows a hard-sharing regime.


\section{\framework{}}
\label{sec:methods}

In this section, we introduce \framework{}, which leverages multiple vision tasks and various data sources to learn a unified representation that transfers well to a wide range of downstream tasks. 
It combines the superior modelling capacity of a split architecture design with the simplicity of hard parameter sharing.
The whole two-stage framework is shown in \cref{fig:stage3-framework}.
In \dilationstage{}, we learn individual sub-backbones for different tasks with multi-source data in parallel. We further interconnect them to an expanded backbone that effectively alleviates interference among tasks. We then condense the expanded backbone to a normal-sized one in \squeezestage{}, producing the final general representation for downstream transfer.

 
 
\subsection{Multi-Task and Multi-Source Learning}
\label{sec:setting}

As illustrated in \cref{fig:figure1} (a), the most common supervised learning setting involves only one task with a single source, i.e., a datum from the source has one label or annotation corresponding to the only task (SSST). There is no task interference during optimization, yet the generated representation is weak in terms of transferability to other tasks.

Traditional multi-task approaches in previous works concurrently learn multiple tasks within a single data source (SSMT), which is shown in \cref{fig:figure1} (b). 
The single data source should have multiple sets of labels, each for one task. Such a data source is hardly scalable due to the high annotation cost.

To fix the drawbacks of previous setups, we propose our multi-source multi-task setting (MSMT), which is displayed in \cref{fig:figure1} (c). More concretely, let $T$ be the number of tasks, then for each task $t\in\{1,2,...,T\}$, there are $N_t$ data sources $\mathcal{S}^t=\{(X_n^t,Y_n^t)\}_{n=1}^{N_t}$ with labels of the task. In this way, we only require $N=\sum_{t=1}^TN_t$ single-task data sources which are easily attainable, avoiding the difficulty of multi-task annotation.
Our setting is also highly extensible since adding new tasks or data sources becomes an effortless process.
During training, the optimization objective of our multi-task and multi-source paradigm is to simply minimize the average loss over all the $N$ data sources consisting of $T$ different tasks:
\begin{align}
    \label{eq:loss}
    \min_\theta L(\theta,\{\mathcal{S}^t\}_{t=1}^T)=\frac{1}{N}\sum_{t=1}^{T}\sum_{n=1}^{N_t}\ell_t(\theta,(X_n^t,Y_n^t))
\end{align}
where $\theta$ denotes model parameters, and $\ell_t$ refers to the loss function for task $t$.





\begin{algorithm}[t]
 \caption{\dilationstage{}}
 \begin{algorithmic}[1]
 \renewcommand{\algorithmicrequire}{\textbf{Input:}}
 \renewcommand{\algorithmicensure}{\textbf{Output:}}
 \REQUIRE 
        Data sources of $T$ tasks $\{\mathcal{S}^t\}_{t=1}^T$, where $\mathcal{S}^t=\{(X_n^t,Y_n^t)\}_{n=1}^{N_t}$; Sub-backbones $\{\mathcal{E}^t\}_{t=1}^T$; Task losses $\{\ell_t\}_{t=1}^T$; Set of reconciliation layers $\gamma$; Total step number $K$; Step threshold $\tau$
 \ENSURE pre-trained expanded backbone $\mathcal{E}$
  \STATE 
  Initialize $\{\mathcal{E}^t\}_{t=1}^T$ and $\gamma$
  \FOR {$k \gets 1 $ to $K$}
  \FOR {$t \gets 1 $ to $T$}
  \STATE Sample a batch $\mathcal{B}^t$ from $\mathcal{S}^t$ with $N_t$ sources
  \IF { $k \leq \tau $}
    \STATE Forward with data $\mathcal{B}^t$ on sub-backbone $\mathcal{E}^t$
    \STATE Compute task loss $\ell_t$
    \STATE Update $\mathcal{E}^t$ separately with gradients from $\ell_t$
  \ENDIF
  \ENDFOR
  \IF { $ k > \tau $ }
  \STATE Forward with multi-task data $\{\mathcal{B}^t\}_{t=1}^T$ on expanded backbone $\{\mathcal{E}^t\}_{t=1}^T\cup\gamma$
  \STATE Compute averaged loss $L$ with \cref{eq:loss}
  \STATE Jointly update $\{\mathcal{E}^t\}_{t=1}^T\cup\gamma$ with gradients from $L$
  \ENDIF
  \ENDFOR
 \RETURN $\{\mathcal{E}^t\}_{t=1}^T\cup\gamma$
 \end{algorithmic} 
 \label{alg:pretrain_dilation}
 \end{algorithm}

\vspace{3mm}
\subsection{\dilationstage{}}
\label{sec:dilation}
 
We aim to learn general representation from heterogeneous tasks while being least affected by the harmful interference among tasks. This motivates us to design this \dilationstage{} to learn a split architecture combining multiple single-task networks. We first train $T$ sub-backbones individually for the $T$ tasks, leveraging their own data sources. 
We then join all $T$ sub-backbones into one holistic architecture, integrating information learned from all tasks to form a general representation. Specifically, we introduce an expanded backbone composed of multiple sub-backbones corresponding to $T$ tasks, along with several \translayer{}s for connecting them, which we describe in detail below. The expanded backbone learned in this pipeline largely 1) preserves the high precision of single-task training,
and 2) combines advantages of all tasks to achieve better generalizability on downstream tasks. The full training process is summarized in \cref{alg:pretrain_dilation}.

\vspace{2mm}
\noindent\textbf{Reconciliation Layer.} 
As shown in \cref{fig:stage3-framework} (a), each \translayer{} is a link between two sub-backbones of two tasks.
It obtains features from one task, transforms them with a few operations, and then fuses them into the features of another task at the same or a deeper layer.

Suppose each sub-backbone has $D$ output layers, and we denote the original output of layer $i\in\{1,2,...,D\}$ from the sub-backbone for task $t\in\{1,2,...,T\}$ by $\mathcal{E}^t_i$.
Let $\gamma_{j \rightarrow i}^{k \rightarrow t}$ ($j\leq i$, $k\neq t$) refer to the \translayer{} taking $\mathcal{E}_j^k$ as input and providing its output to the $i^\text{th}$ layer of another task $t$. According to \cref{fig:stage3-framework} (a), $\gamma_{j \rightarrow i}^{k \rightarrow t}$ can be expressed as the composition of one $\gamma_b$ and $i-j$ times of $\gamma_a$. 
Receiving all cross-task and cross-layer features, we take a summation to compute the final fused output $F^t_i$ at layer $i$ of the sub-backbone for task $t$:
\begin{align}
    F^t_i = \mathcal{E}^t_i+\sum_{\substack{k=1\\k\neq t}}^T\sum_{j=1}^i\gamma_{j \rightarrow i}^{k \rightarrow t}\left(\mathcal{E}^k_j\right).
\end{align}
%

Adding \translayer{}s directly facilitates interactions among information from different tasks. Thus it closely unifies all sub-backbones into one expanded backbone expressing an integrated and general representation. In practical implementation, to avoid task interference introduced by such cross-task communication, we detach inputs to all \translayer{}s from the computational graph to cut off further gradient propagation.

\begin{figure*}[t]
\centering
 \includegraphics[width=1.0\linewidth]{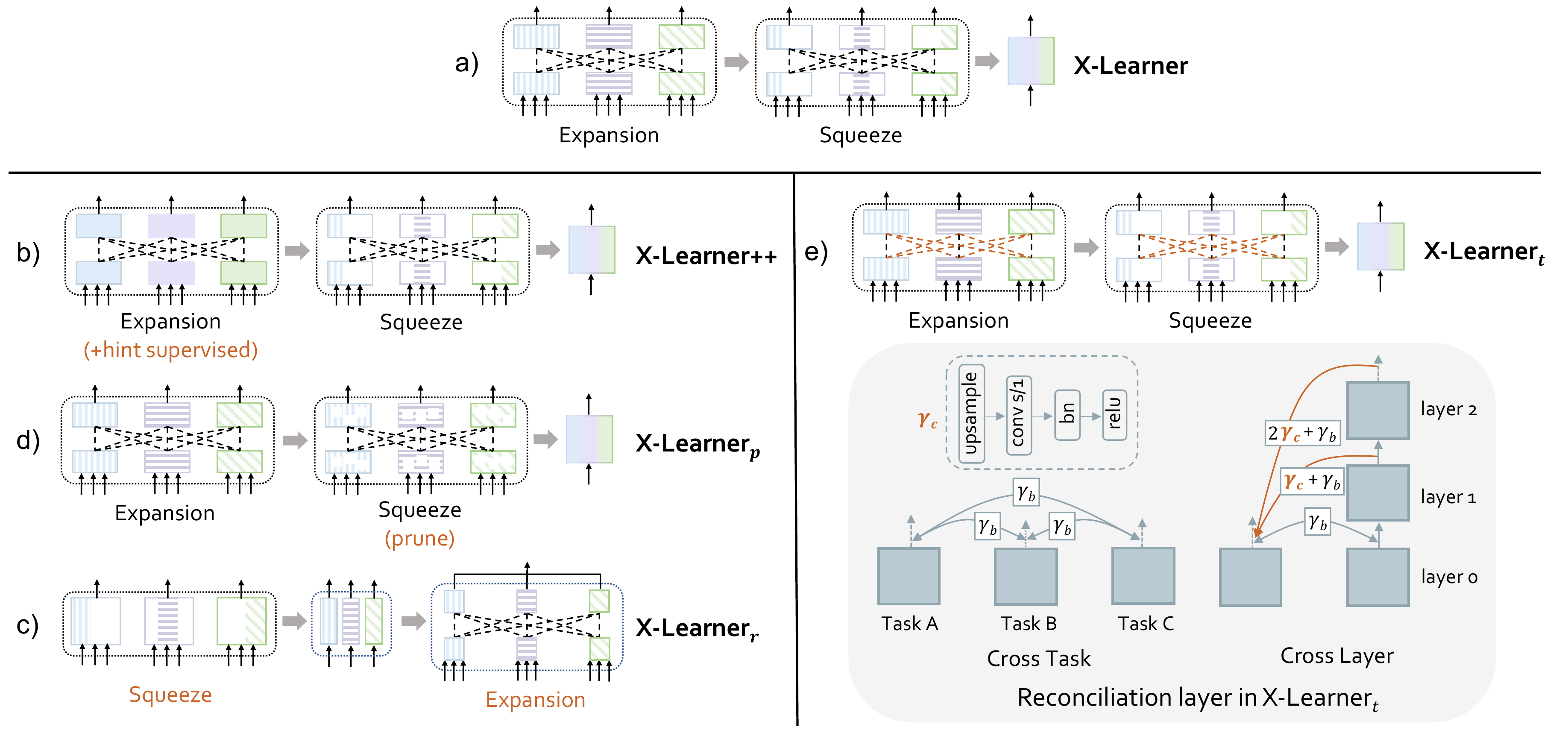}
 \caption{\textbf{Variants of \framework{}.} (a) is the default form of \textbf{X-Learner}. (b) The expansion stage of \textbf{X-Learner++} is supervised by extra hints from single-task single-source pre-trained models. (c) X-Learner$_r$ is a Squeeze-Expansion version. (d) X-Learner$_p$ replace the distillation with pruning in the squeeze stage. (e) We switch to a new reconciliation layer in X-Learner$_t$. Differences between variants and the default X-Learner are highlighted in red.
 }
 \label{fig:stage3-experiment-module}
 \vspace{-2mm}
\end{figure*}

\subsection{\squeezestage{}}
\label{sec:squeeze}

The previous \dilationstage{} gives a concerted representation provided by the expanded backbone uniting all $T$ sub-backbones of $T$ tasks.
However, it also introduces an undesirable $T$ times increase in the number of model parameters and computational complexity. To maintain performance while reducing the expanded parameters, we present the \squeezestage{}. The final squeezed model remains highly generalizable for downstream transfer while sharing the same number of parameters with a single-task sub-backbone.

In \squeezestage{}, given an expanded backbone, we adopt distillation to consolidate the model.
We employ the FitNets~\cite{romero2014fitnets} approach, but with multiple targets (hints) from the expanded backbone as the student's supervision. Formally, given multiple outputs from the expanded teacher indexed by $t\in\{1, 2, ..., T\}$,
we refer to $F^{t}$ as the output feature of task $t$, and $\hat{F}$ as the feature of the student network. We perform distillation between the student model and the bunch of teacher outputs. Specifically, we project the single student feature $\hat{F}$ through a task-specific guidance layer $\mathcal{G}^{t}$, and expect the outcome to match the teacher's version $F^{t}$. Therefore, our distillation loss $L_\text{squeeze}$ is simply the sum over squared $L_2$ losses of all teacher-student pairs:
\begin{align}
    L_\text{squeeze} = \sum_{t=1}^T{ \norm\Big{{F^{t} -  \mathcal{G}^{t}(\hat{F})}}^2_2 }
\end{align}
The guidance layer $ \mathcal{G}^{t}$  is composed of a convolutional layer and a normalization layer:
\begin{align}
    \mathcal{G}^{t}(x) = \text{Norm}(\text{Conv}(x)).
\end{align}
We adopt an $1\times1$ convolution which transforms the student's feature to have the same number of channels as the teacher's output.
For the normalization function, we simply choose Batch Normalization~\cite{ioffe2015batch} as in \cite{romero2014fitnets}.


\subsection{Variants of \framework{}}
\label{sec:variants}


\framework{} is a highly flexible multi-task pre-training framework, and many variants can be designed from the default setting. In this section, we describe several possibilities, which are illustrated in \cref{fig:stage3-experiment-module}. More detailed differences among those variants are listed in \cref{fig:stage3-params}.


\vspace{2mm}
\noindent{\textbf{\framework{}$_{\boldsymbol{r}}$}}.
We notice that the number of parameters in each individual model is first rising and then declining in our default \framework{}. It is natural to also study the reversed order, i.e., Squeeze-Expansion. In the new squeeze stage, we use $T$ task-specific teachers trained with multiple sources to distill $T$ more light-weight sub-backbones. They are then combined into one network with normal computational complexity via \translayer{}s in the following expansion stage.

\vspace{2mm}
\noindent{\textbf{\framework{}$_{\boldsymbol{t}}$}}. We make a modification on the \translayer{}s and let them take features from deeper layers of other sub-backbones as input and fuse to low-level features of a task. We also replace $\gamma_a$ in cross-layer reconciliation layers with $\gamma_c$ which is composed of an up-sampling layer and a convolutional layer. 

\vspace{2mm}
\noindent{\textbf{\framework{}$_{\boldsymbol{p}}$}}. We replace the distillation operation with unstructured pruning in \squeezestage{}. It is another way to reduce  computation consumption while maintaining the performance of a network. We adopt a simple unstructured pruning method referencing \cite{PRUNE}.


\vspace{2mm}
\noindent{\textbf{\framework{}++}}. Inspired by~\cite{li2021representation}, in the Expansion Stage, we add extra supervisions from single-task single-source pre-trained model in the form of hints besides the original supervision from labels of multiple data sources. This can be viewed as adding a pre-distillation process with multiple SSST teachers prior to training the expanded backbone.






\begin{table}[t]
\centering
\footnotesize
\caption{\textbf{Datasets used for \framework{} pre-training.} We grouped them into manually defined image domains according to~\cite{mensink2021factors}.} 
\label{tab:pre-train dataset}
\begin{tabular}{llll}
\toprule
Dataset & Task &  Domain &  Train Size \\ \hline
ImageNet~\cite{russakovsky2015imagenet}& General CLS.& Websearch & 1.3M  \\
Places365~\cite{zhou2017places}& General CLS. & Websearch & 8.0M \\
iNat2021~\cite{van2021benchmarking}& Fine-Grained CLS. & Consumer & 2.7M \\
CompCars~\cite{yang2015large}& Fine-Grained CLS.& Close-ups & 120k \\
Tsinghua Dogs~\cite{zou2020new}& Fine-Grained CLS.& Close-ups & 65k \\ \hline
COCO~\cite{lin2014microsoft}& General DET. & Consumer & 118k \\
Objects365~\cite{shao2019objects365}& General DET. & Consumer & 609k \\
WIDER FACE~\cite{yang2016wider}& Face DET. & Websearch & 13k \\
\hline
ADE20K~\cite{zhou2019semantic}& Semantic SEG. & Consumer & 20k \\
COCO-Stuff~\cite{caesar2018cocostuff} & Semantic SEG. & Consumer & 164k \\
\bottomrule
\end{tabular}
\end{table}
\begin{table}
\centering
\caption{\textbf{Comparison with supervised and self-supervised methods on classification, detection and segmentation.} \textcolor{red}{*} represents the model is not pre-trained with semantic segmentation. We compare X-Learner to supervised pre-training, self-supervised learning, and a simple hard-sharing multi-task learning baseline. Relative gains are computed with respect to the ImageNet supervised baseline.\label{tab:tab_main_results}}
\resizebox{1.0\linewidth}{!}
{
\begin{tabular}{lccc}
\toprule
\multicolumn{1}{l|}{Method}& AVG Cls               & PASCAL Det           & PASCAL Seg           \\ \hline
\multicolumn{1}{l|}{ImageNet~\cite{russakovsky2015imagenet} Supervised} & 74.4& 81.5& 75.7\textcolor{red}{*}\\
\multicolumn{1}{l|}{SimCLR~\cite{chen2020simple}}  & 74.6& 82.9& 74.1\textcolor{red}{*}\\
\multicolumn{1}{l|}{Hard-sharing}  & 73.2& 83.7& 70.5\textcolor{red}{*}\\ \hline
\multicolumn{1}{l|}{X-Learner} & 77.1 \textcolor{blue}{(+2.7)}&  84.4 \textcolor{blue}{(+2.9)}& 77.1\textcolor{red}{*} \textcolor{blue}{(+1.4)}       \\ 

\multicolumn{1}{l|}{X-Learner++} & 77.4 \textcolor{blue}{(+3.0)}& \textbf{84.8} \textcolor{blue}{(+3.3)}& \textbf{77.5}\textcolor{red}{*} \textcolor{blue}{(+1.8)}\\ \hline
\multicolumn{1}{l|}{X-Learner w/ seg}   & \textbf{77.7} \textcolor{blue}{(+3.3)}& 84.3 \textcolor{blue}{(+2.8)}& \textbf{77.6} \textcolor{blue}{(+1.9)}\\  
\bottomrule
\end{tabular}
}
\vspace{-4mm}
\end{table}

\section{Experiments}
\label{sec:exps}

\subsection{Pre-Training Settings}
\label{sec:exp-setting}

\noindent\textbf{Pre-Training Sources (Datasets).}
\cref{tab:pre-train dataset} summarizes the sources we use for experiments. 
Most of our experiments are conducted in a base setting, where we pre-train models with 2 tasks: classification and object detection.
We use 3 sources for image classification: ImageNet~\cite{russakovsky2015imagenet}, iNat2021~\cite{van2021benchmarking} and Places365~\cite{zhou2017places} (Challenge version), and 2 sources for object detection: COCO~\cite{lin2014microsoft} and Objects365~\cite{shao2019objects365}. We also consider two extended settings: 1) to investigate the effect of more sources on \framework{}, we add CompCars~\cite{yang2015large} as well as Tsinghua Dogs~\cite{zou2020new} as two extra classification sources, and select WIDER FACE~\cite{yang2016wider} as a new object detection source; 2) we study the impact of adding a new task, which is semantic segmentation, with ADE20K~\cite{zhou2019semantic} and COCO-Stuff~\cite{caesar2018cocostuff} as its sources.

\vspace{2mm}
\noindent\textbf{Implementation Details.}
We implement \framework{} and its variants described in \cref{sec:variants} using ResNet-50~\cite{he2016deep} as the basic backbone throughout our experiments unless otherwise specified.
\begin{figure*}[htb]
\footnotesize
\begin{minipage}[htb]{0.6\linewidth}
\centering
\resizebox{1.0\linewidth}{!}
{
\begin{tabular}{l|c|ccc|c}
\toprule
\multicolumn{1}{l|}{Experiment} & \multicolumn{1}{l|}{Sub-Backbone} & \multicolumn{1}{l}{Expansion} & \multicolumn{1}{l}{Squeeze} & \multicolumn{1}{l|}{Pre-Distillation} & \multicolumn{1}{l}{Parameters}         \\ \hline
Hard-sharing      & ResNet-50  & $\times $& $\times $& $\times $& $\to$    \\
\framework{} & ResNet-50  & $\surd $& D & $\times $& $\nearrow$ $\searrow$ \\

\framework{}$_{r}$& HalfResNet-50& $\surd $& D & $\times $     & $\searrow$ $\nearrow$ \\

\framework{}$_{t}$& ResNet-50  & $\surd $& D & $\times $& $\nearrow$ $\searrow$ \\

\framework{}$_{p}$& ResNet-50  & $\surd $& P & $\times $& $\nearrow$ $\searrow$ \\

\framework{}++ & ResNet-50  & $\surd $& D & $\surd $& $\nearrow$ $\searrow$ \\ 


\framework{} w/o Rec. & ResNet-50  & $\times $& D & $\times $&  $\searrow$ \\ \bottomrule
\end{tabular}}
\end{minipage}
\begin{minipage}[htb]{0.45\linewidth}  
\centering
\includegraphics[width=0.8\linewidth]{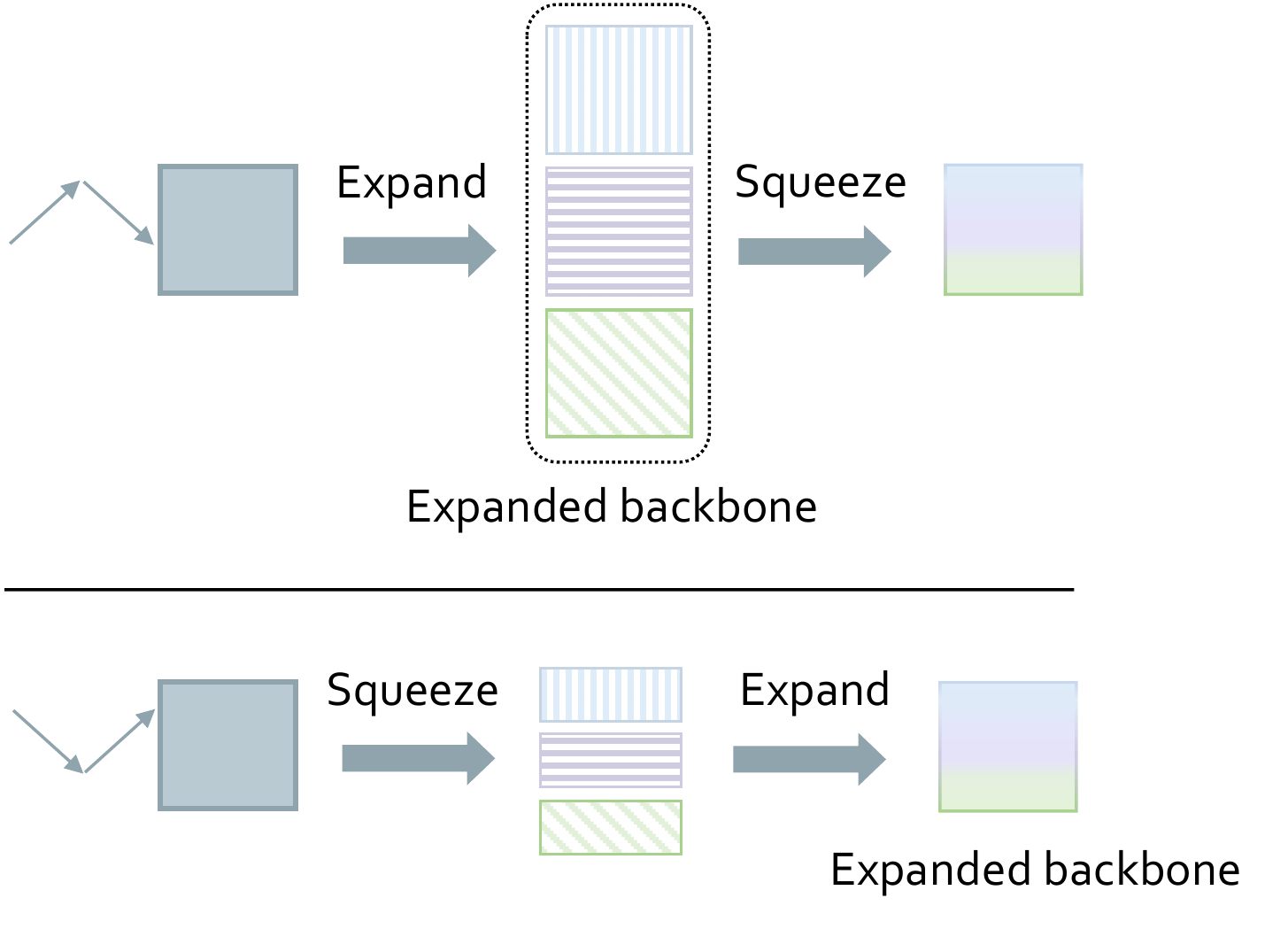}
\end{minipage}
\caption{\textbf{Differences among X-learner variants.} We conduct different ablation study of X-Learner. Pre-distillation refers to applying extra supervisions from single-task single-source pre-trained models as is introduced in \framework{}++. In the Squeeze column, we denote distillation by D and pruning by P if there is a squeeze stage present in the pipeline. The change of the parameter can refer to the figure on the right.
\label{fig:stage3-params}}
\end{figure*}
The weights of \translayer{}s are initialized with \cite{glorot2010understanding}. We use SGD optimizer with a momentum of 0.9~\cite{sutskever2013importance}, $10^{-4}$ weight decay and a base learning rate of 0.2. We decay the learning rate three times by a multi-step schedule with factors 0.5, 0.2 and 0.1 at 50\%, 70\% and 90\% of the total iterations respectively.

\subsection{Downstream Task Settings}
\label{sec:eval_protocol}
\noindent\textbf{Classification.}
We select 10 datasets from the well-studied evaluation suite introduced by \cite{supervised1}, including general object classification (CIFAR-10\cite{krizhevsky2009learning}, CIFAR-100\cite{krizhevsky2009learning}); fine-grained object classification (Food-101\cite{downfood}, Stanford Cars\cite{downcar}, FGVC-Aircraft\cite{downaricraft}, Oxford-IIIT Pets\cite{downpets}, Oxford 102 Flower\cite{downflowers}, Caltech-101\cite{downcaltech}), and scene classification (SUN397\cite{downsun}). We follow the linear probe evaluation setting used in \cite{clip}. We use the average accuracy of 10 classification datasets (AVG Cls) to represent the overall performance on the classification task. We train a logistic regression classifier using the L-BFGS optimizer, with a maximum of $1,000$ iterations. We search the value for the L2 regularization strength $\lambda$ over a set which distributes evenly over the range between $10^{-1}$ and $10^{-5}$. We use images of resolution $224 \times 224$ for both training and evaluation.

\vspace{2mm}
\noindent \textbf{Detection.}
We fine-tune our pre-trained model on PASCAL VOC07+12 (PASCAL Det)~\cite{Everingham10VOC} for the detection task. We use Faster-RCNN~\cite{fasterrcnn} architecture in our experiments and run 24,000 iterations with a batch size of 16. We use SGD as the optimizer and search the best learning rate between 0.001 and 0.05. Weight decay is set to $10^{-4}$, and momentum is set to 0.9. Evaluation is performed on the PASCAL VOC 2007 test set, with the shorter edges of images scaled to 800 pixels.

\vspace{2mm}
\noindent \textbf{Semantic Segmentation.}
We evaluate models on PASCAL VOC 2012 (PASCAL Seg)~\cite{Everingham10VOC}. We run 33,000 iterations with a batch size of 16. The architecture is based on Deeplab v3~\cite{deeplab}. We use SGD as the optimizer with a learning rate between 0.001 and 0.07. Weight decay is set to $10^{-4}$, and momentum is set to 0.9. Images are scaled to $513\times513$.
%



\setlength{\tabcolsep}{4pt}
\begin{table*}[t]
\centering
\caption{\textbf{Comparison on extended settings with extra pre-training sources.} 
By adding sources in different tasks (marked in bold italic), Hard-sharing suffers performance drops on both upstream and downstream tasks, while our \framework{} is stable across different settings, benefiting from the proposed \textit{Expansion Stage}.
}
\large
\resizebox{16cm}{!}{
\begin{tabular}{lc|cccccccc|cc}
\toprule
\multicolumn{2}{c|}{}    & \multicolumn{8}{c|}{Pre-train}   & \multicolumn{2}{c}{Transfer} \\ \hline
\multicolumn{1}{l|}{Expriments}  & Methods   & ImageNet & iNat2021 & Places & \textbf{\textit{Cars}} & \multicolumn{1}{c|}{\textbf{\textit{Dogs}}} & COCO & Objects365 & \textbf{{\textbf{\textit{FACE}}}} & \multicolumn{1}{c|}{AVG Cls} & PASCAL Det \\ \hline
\multicolumn{1}{l|}{\multirow{2}{*}{Base }} & Hard-sharing & \multicolumn{1}{c}{75.0}  & \multicolumn{1}{c}{75.3} & \multicolumn{1}{c}{53.0}  & {--} & \multicolumn{1}{c|}{{--}}    & \multicolumn{1}{c}{35.5} & \multicolumn{1}{c}{17.4} & {--}   & \multicolumn{1}{c|}{73.2}    & 83.7    \\
\multicolumn{1}{c|}{}& X-Learner & 77.3& 79.7   & 54.4   & {--} & \multicolumn{1}{c|}{{--}}           & 39.9 & 22.2      & {--}      & \multicolumn{1}{c|}{77.1}    & 84.4    \\ \hline
\multicolumn{1}{l|}{\multirow{2}{*}{$+$ Cls Sources}}            & Hard-sharing & 73.7     & 73.6        & 52.3   & 98.5     & \multicolumn{1}{c|}{85.3}         & 35.4 & 17.6      & {--}        & \multicolumn{1}{c|}{77.5}    & 83.1    \\
\multicolumn{1}{c|}{}                                                       & X-Learner & 77.3     & 77.9        & 54.4   & 98.4     & \multicolumn{1}{c|}{86.9}         & 40.5 & 22.6      & {--}        & \multicolumn{1}{c|}{80.6}    & 84.3    \\ \hline
\multicolumn{1}{l|}{\multirow{2}{*}{$+$ Cls \& Det Sources}}   & Hard-sharing & 73.6     & 73.6        & 52.0   & 98.4     & \multicolumn{1}{c|}{85.4}         & 34.9 & 16.5      & 31.5      & \multicolumn{1}{c|}{77.1}    & 83.2    \\
\multicolumn{1}{c|}{}                                                      & X-Learner & 76.9     & 78.6        & 54.6   & 98.6     & \multicolumn{1}{c|}{85.9}         & 40.1 & 22.1      & 33.6      & \multicolumn{1}{c|}{80.5}    & 84.3    \\ \bottomrule
\end{tabular} }
\label{tab:add-dataset}
\end{table*}
\subsection{Main Results}
\label{sec:main_result}

\begin{table*}
\small
\centering
\caption{\textbf{Comparison with self-training.} 
%
%
PASCAL Seg is an unseen task for X-Learner$++$, which is marked with~{\color{red}*}. NYU-Depth V2 is an unseen task for X-Learner$_{R152}$, which is marked with~{\color{red}*}. }
\label{tab:must}
\resizebox{1.0\linewidth}{!}
{
\begin{tabular}{lc|c|cccc}
\toprule
\multicolumn{1}{l|}{Method}& Backbone & Pre-training Settings & CIFAR-100~\cite{krizhevsky2009learning}     & PASCAL Det~\cite{Everingham10VOC} & PASCAL Seg~\cite{Everingham10VOC}   & NYU-Depth V2~\cite{silberman2012indoor}      \\ \hline
\multicolumn{1}{l|}{MuST~\cite{ghiasi2021multi}}  & ResNet-152 & ImageNet + DET. + SEG. + DEP. & 86.3& 85.1& 80.6 & 87.8\\ 
\multicolumn{1}{l|}{MuST~\cite{ghiasi2021multi}}  & ResNet-152 & JFT300M + DET. + SEG. + DEP. & 88.3& 87.9& \textbf{82.9} & 89.5\\ 
\multicolumn{1}{l|}{X-Learner$++$} & ResNet-50 & ImageNet + DET.  & 87.0  \textcolor{blue}{(+0.7)}& 87.3~\textcolor{blue}{(+2.2)}& 78.8\textcolor{red}{*} \textcolor{red}{(-1.8)} & 89.0~\textcolor{blue}{(+1.2)} \\
\multicolumn{1}{l|}{X-Learner$_{R152}$}  & ResNet-152 & ImageNet + OBJ365 + COCO  & \textbf{88.7} \textcolor{blue}{(+2.4)}& \textbf{88.5} \textcolor{blue}{(+3.4)}& \textbf{81.4} \textcolor{blue}{(+0.8)} & \textbf{91.2}\textcolor{red}{*}\textcolor{blue}{(+3.4)} \\
\multicolumn{1}{l|}{X-Learner$_{R152}$}  & ResNet-152 & ImageNet + DET. + SEG.   & \textbf{89.7} \textcolor{blue}{(+3.4)}& \textbf{88.6} \textcolor{blue}{(+3.5)}& \textbf{82.6} \textcolor{blue}{(+2.0)} & \textbf{91.3}\textcolor{red}{*}\textcolor{blue}{(+3.5)} \\
\bottomrule
\end{tabular}
}
\vspace{-4mm}
\end{table*}
\begin{table*}[ht]
\footnotesize
\centering
\caption{\textbf{The effect of applying reconciliation layers in the Expand Stage.} The reconciliation layer can significantly improve the performance in multi-task learning.\label{tab:ablation_of_recon}}
\begin{tabular}{lcc} 
\toprule
 &  \small{~~~AVG Cls~~~}   & \small{~~~PASCAL Det~~~~} \\ 
\midrule[1pt]
\small{X-Learner w/o Rec}& \small{74.8} & \small{83.9}  \\ %
\small{\framework{}}  & \small{\textbf{77.1}}& \small{\textbf{84.4}}    \\
\bottomrule
\end{tabular}
\end{table*}

\begin{table*}[ht]
\centering
\small
\caption{\textbf{Comparison of various X-Learner variants.} Pre-training tasks and downstream tasks are evaluated on X-Learner variants. Our framework always performs better than Hard-sharing.}
\begin{tabular}{l|ccccc|cc}
\toprule
\multicolumn{1}{l|}{} & \multicolumn{5}{c|}{Pre-train}& \multicolumn{2}{c}{Transfer}  \\ \cline{2-8} 
Method & ImageNet   & iNat2021 & Places & COCO  & Objects365 & AVG Cls   & PASCAL Det \\ \hline

Hard-sharing& \multicolumn{1}{c}{75.0} & \multicolumn{1}{c}{75.3} & \multicolumn{1}{c}{53.0} & \multicolumn{1}{c}{35.5} & 17.4 & \multicolumn{1}{c}{73.2}& 83.7    \\
\framework{}& \multicolumn{1}{c}{77.3} & \multicolumn{1}{c}{79.7} & \multicolumn{1}{c}{54.4} & \multicolumn{1}{c}{39.9} & 22.2 & \multicolumn{1}{c}{77.1}& 84.4    \\


%
\framework{}$_{r}$& \multicolumn{1}{c}{73.9} & \multicolumn{1}{c}{76.6} & \multicolumn{1}{c}{52.5} & \multicolumn{1}{c}{41.1} & \multicolumn{1}{c|}{21.7}   & \multicolumn{1}{c}{73.9}   & 84.1    \\
%

%
X-Learner$_{t}$& \multicolumn{1}{c}{76.3} & \multicolumn{1}{c}{79.9} & \multicolumn{1}{c}{53.3} & \multicolumn{1}{c}{42.5} & 22.0& \multicolumn{1}{c}{74.5}& 83.5    \\

\framework{}$_{p}$  & \multicolumn{1}{c}{76.1} & \multicolumn{1}{c}{78.6} & \multicolumn{1}{c}{53.5} & \multicolumn{1}{c}{42.4} & 23.4 & \multicolumn{1}{c}{77.2}& 83.1    \\
%

\framework{}++& \multicolumn{1}{c}{77.2} & \multicolumn{1}{c}{80.4} & \multicolumn{1}{c}{54.6} & \multicolumn{1}{c}{40.1} & 22.4 & \multicolumn{1}{c}{\textbf{77.4}} &  \textbf{84.8}    \\
\bottomrule
\end{tabular}
\label{tab:main_result}
\end{table*}

\noindent \textbf {Pre-Training Paradigm Comparison.}~\cref{tab:tab_main_results} compares our pre-training scheme \framework{} with supervised training and self-supervised learning (SimCLR~\cite{chen2020simple}) on ImageNet~\cite{russakovsky2015imagenet}, as well as a simple hard-parameter-sharing baseline (named as ``Hard-sharing'') on our multi-task and multi-source setting. We report performances on all three types of downstream tasks.
Under the base setting, \framework{} uniformly outperforms all compared methods in terms of all evaluated metrics, especially AVG Cls. We also observe that the Hard-sharing model has better performance than the ImageNet-supervised model on PASCAL Det, but suffers a performance drop of 1.2\% in AVG Cls. This suggests that the hard-sharing model benefits from multi-task pre-training with object detection sources included, but is harmed by task interference. In contrast, our \framework{} clearly overcomes the shortcoming and alleviates undesirable interference, leading to performance boosts on all considered tasks. Moreover, compared with training solely on ImageNet which is already specialized for classification, our approach still enjoys a 2.5\% increase on AVG Cls. This result demonstrates that our setting of learning with multiple tasks simultaneously is beneficial for all involved pre-training tasks, such as classification here.

In addition, our \framework{}++ mentioned in \cref{sec:variants} further enhances performance by means of its extra distillation process during sub-backbone training in the Expansion Stage, and achieves the best performance on all three downstream tasks.
 
We also compare our \framework{}++ with the multi-task self-training method MuST~\cite{ghiasi2021multi} in \cref{tab:must}, For fair comparison, we fine-tune on the CIFAR-100 dataset instead of applying our default linear probe setting, evaluate PASCAL Det with pre-trained FPN~\cite{lin2017feature}, and set output stride to 8 in segmentation. 

Our model surpasses MuST on classification and detection tasks despite using ResNet-50 instead of the more advanced ResNet-152 applied by MuST.
To better show the effectiveness of our setting, we also conduct an experiment with the ResNet-152 backbone. \cref{tab:must} shows the performance of X-Learner$_\text{R152}$ as well as MuST on four different tasks. We observe that our framework outperforms the self-training method by significant margins on all evaluated downstream tasks.
Moreover, it is worth mentioning that on NYU-Depth V2, our X-Learner, without any depth estimation pre-training, surpasses MuST which is learned with MiDaS, a  mixture dataset with 10 depth-wise datasets. This zero-shot result further demonstrates the strong generalization capability of X-Learner.

We also compare our X-Learner$_{R152}$ with a stronger version of MuST model pre-trained with JFT-300M, which is much larger than our datasets. As our X-Learner achieves 89.7 and 88.6 in downstream classification and detection tasks. This comparison proves that the dataset size is not an important factor, and our design has its superiority.

 %

 

\vspace{2mm}
\noindent \textbf{Cross-Task Generalization and Scalability.}
In \cref{tab:tab_main_results}, among methods that are not pre-trained on semantic segmentation, our \framework{}++ has the highest result on PASCAL Seg. This validates that our models produce more generalizable representations in terms of unseen tasks.

In addition to generalizability, our framework is also highly scalable and can incorporate extra tasks or sources effortlessly. As a demonstration, we add a semantic segmentation task according to the extended setting with ADE20K and COCO-Stuff. Results of ``\framework{} w/ seg'' in \cref{tab:tab_main_results} show improvement on PASCAL Seg by 0.5 mIoU compared to the basic \framework. Classification performance is also benefitted from the new task introduced, demonstrating the effectiveness of our multi-task learning approach.

%

\noindent \textbf{Necessity of Reconciliation Layers.}
As shown in~\cref{tab:ablation_of_recon}, we train an X-Learner without \translayer{} to study the importance of the component. Compared to the default setting, removing \translayer{}s leads to significant performance drops at downstream transfer learning, especially on fine-grained datasets. We find that the feature from detection sub-backbone contains more detail, and it can be enhanced to a universal feature by the reconciliation layer. This phenomenon also verifies that \translayer{}s play a crucial role in coordinating multiple tasks towards the common goal of general representation learning.

\subsection{In-Depth Studies}


\subsubsection{Multi-Task and Multi-Source Pre-Training}
~\

\noindent \textit{Observation 1: Proper multi-task learning promotes collaboration instead of bringing interference.}
%
As is discussed in \cref{sec:main_result}, \framework{} not only resolves the task interference issue encountered by the hard-sharing model, but also surpasses single-task pre-trained models such as the ImageNet baseline in terms of downstream results. This shows that with an appropriately designed learning scheme, multi-task training is able to collaboratively enhance performances on all pre-training tasks. This conclusion is again corroborated by the results of \framework{}++ in \cref{tab:tab_main_results}. With a more elaborated design, performances on all tasks are again consistently boosted.
%
%

 
\vspace{+3mm}
\noindent \textit{Observation 2: Additional sources further improve multi-task and multi-source representation learning if task conflicts are well-mitigated.}
%
We experiment on the extended setting with extra classification and detection sources. 
%
%
The added sources, such as CompCars~\cite{yang2015large}
and WIDER FACE~\cite{yang2016wider}, have data in domains very different from existing sources. Ideally, including sources of complementary nature should help the overall multi-task and multi-source learning, since information available for pre-training is enriched and is more likely to cover downstream domains. However, this may also increase conflicts among tasks if not dealt with properly. In \cref{tab:add-dataset}, we can see that the over-simplified hard-sharing baseline has considerably inferior results at both upstream and downsteam if more sources are added. In pre-training stage, there is slight decrease after adding classification sources. This is due to the increase in task conflict when introducing new data domains. Nonetheless, we can find that additional sources becomes beneficial to transfer learning tasks both in hard-sharing and X-Learner. Compared to hard-sharing, X-Learner has mitigated such detrimental conflict to a certain extent with the aid of our two-stage design. 
This suggests that when task interference is properly alleviated, new data sources can be fully utilized by the model to learn more diverse knowledge and enhance the final representation.
%


\subsubsection{Design of \framework{} Framework}
~\

\noindent \textit{Observation 3: Expansion-Squeeze is better than Squeeze-Expansion. }
%
In \cref{sec:variants}, we have described the \framework{}$_{r}$ variant in which the order of the two stages within \framework{} is reversed. Performing squeezing first would result in smaller single-task sub-backbones with $1/T$ of the original size. Since $T=2$ in our base setting, we should get two halved ResNet-50 models, corresponding to HalfResNet-50 in \cref{fig:stage3-params}, which are to be joined in the further expansion process. HalfResNet-50 is a sub-backbone with only $1/\sqrt{2}$ of the original ResNet-50 channels.
As shown in \cref{tab:main_result}, \framework{}$_{r}$ has lower performance on most pre-training tasks and all downstream tasks than the default \framework{}.
This finding is reasonable since by intuition, shrinking sub-backbones first is likely to cause unrecoverable information loss.
It also validates our choice of Expansion-Squeeze for the default setup.
Note that \framework{}$_{r}$ is still better than the hard-sharing model, which again highlights the importance of a two-stage paradigm to mitigate task interference.
%
%

\vspace{+2mm}
\noindent \textit{Observation 4: Reconciliation layers should receive information from lower levels.}
We also evaluate the alternative design of \framework{}$_t$, where \translayer{}s take features from deeper layers instead of shallower ones.
Experiments in \cref{tab:main_result} show that the modified and original setups are both competitive at upstream pre-training.
%
However, \framework{}$_t$ is not as good as \framework{} in terms of downstream tasks. 
In conclusion, low-level features are more suitable to serve as complementary information among heterogeneous tasks. 

\vspace{+2mm}
\noindent \textit{Observation 5: Pruning may replace distillation in Squeeze Stage.}
%
In \cref{tab:main_result}, \framework{}$_{p}$ achieves results similar to those of \framework{}. This shows that pruning is also a valid choice for squeezing the expanded backbone, and thus is able to substitute distillation in Squeeze Stage.

\vspace{-2mm}
\section{Discussion and Conclusion}
\label{sec:conclusion}
\vspace{-2mm}
In this paper, we propose a flexible multi-task and multi-source pre-training paradigm called \framework{}, the general framework for representation learning by supervised multi-task learning.  Heterogeneous tasks and diverse sources can be jointly learned with the help of the Expansion Stage and Squeeze Stage. We validate that \framework{} mitigates the well-known task interference problem  and  learns  unified  general  representation  that generalizes  well to multiple seen and unseen tasks. We also show that \framework{} is superior to  traditional supervised and self-supervised learning methods, as well as self-training approaches. 
In addition, We also demonstrate that our framework is highly flexible and additional tasks or sources can be integrated  in a ``plug-and-play'' manner. Moreover, we offer several insightful observations through our experiments.
One possible limitation is that the representation capability of our current pre-training is confined by the scale of publicly available datasets. It is possible to study with larger sources and more tasks in our framework.
We hope this work will encourage further researches towards creating general representations by performing multi-task and multi-source learning at scale.
%

%

{\small
\bibliographystyle{ieee_fullname}
\bibliography{egbib}
}

\end{document}